\documentclass[runningheads]{llncs}
\usepackage[T1]{fontenc}
\usepackage{makecell} 
\usepackage{float}
\usepackage{placeins}
\usepackage{dblfloatfix}

\usepackage{amssymb}      
\usepackage{algorithm}    
\usepackage{algpseudocode}
\algrenewcommand\alglinenumber[1]{\scriptsize #1} 
\usepackage{graphicx}

\makeatletter
\renewcommand{\fnum@algorithm}{\scriptsize Algorithm~\thealgorithm}
\makeatother

\usepackage{listings}

\usepackage{subcaption}   

\usepackage{paralist}
\usepackage{booktabs}
\usepackage{amsmath}
\usepackage{accessibility}
\usepackage{xurl} 
\usepackage{hyperref} 

\usepackage{enumitem}

\usepackage[most]{tcolorbox}

\newtcolorbox{promptbox}{
  enhanced,
  breakable,
  enhanced jigsaw,          
  colback=gray!10,
  colframe=gray!50,
  boxrule=0.25pt,           
  arc=0.5pt,                
  left=1.5pt,               
  right=1.5pt,
  top=1pt,                   
  bottom=1pt,
  fontupper=\ttfamily\scriptsize,  
  width=\textwidth,
  before skip=1.5pt,
  after skip=1.5pt,
}

\usepackage{xcolor}
\usepackage{tcolorbox}
\usepackage{listings}
\tcbuselibrary{listings,skins,breakable}

\lstdefinelanguage{ABAP}{
  morekeywords={
    PERFORM, USING, CALL, TRANSACTION, WITH, AUTHORITY-CHECK, MODE,
    UPDATE, SET, PARAMETER, FIELD, AND, SKIP, FIRST, SCREEN,
    DATA, WRITE, IF, ENDIF, LOOP, ENDLOOP, FORM, ENDFORM,
    BEGIN, END, SELECT, FROM, WHERE, INTO, TABLE, CLEAR, MOVE
  },
  sensitive=true,
  morecomment=[l][\color{gray!60}\itshape]{"},  
  morestring=[b]',                               
}

\newtcblisting{codebox}[1][]{
  listing engine=listings,
  enhanced,
  breakable,
  colback=gray!4,
  colframe=gray!40,
  boxrule=0.25pt,       
  arc=0.5pt,            
  outer arc=0.5pt,
  left=1pt,              
  right=1pt,
  top=1pt,               
  bottom=1pt,
  listing only,
  before skip=2pt,       
  after skip=2pt,
  listing options={
    language=ABAP,
    basicstyle=\ttfamily\scriptsize,  
    keywordstyle=\color{blue!60!black}\bfseries,
    commentstyle=\color{gray!60}\itshape,
    stringstyle=\color{red!50!black},
    showstringspaces=false,
    breaklines=true,
    breakatwhitespace=true,
    keepspaces=true,
    columns=fullflexible,
    aboveskip=0pt,
    belowskip=0pt,
  },
  #1
}

\begin{document}
\title{Towards Practical GraphRAG: Efficient Knowledge Graph Construction and Hybrid Retrieval at Scale}
%
%
\author{
Congmin Min\inst{1} \and
Sahil Bansal\inst{1}\thanks{Sahil Bansal, Joyce Pan, and Abbas Keshavarzi contributed equally to this work.} \and
Joyce Pan\inst{1} \and
Abbas Keshavarzi\inst{1} \and
Rhea Mathew\inst{1} \and
Amar Viswanathan Kannan\inst{1}
}

\authorrunning{C. Min et al.}

\institute{
SAP, Palo Alto, CA, USA \\
\email{\{congmin.min, sahil.bansal01, joyce.pan01, abbas.keshavarzi, rhea.mathew\}@sap.com, amarviswanathan@gmail.com}
}

\maketitle
\begin{abstract}

We propose a scalable and cost-efficient framework for deploying Graph-based Retrieval-Augmented Generation (GraphRAG) in enterprise environments. While GraphRAG has shown promise for multi-hop reasoning and structured retrieval, its adoption has been limited due to reliance on expensive large language model (LLM)-based extraction and complex traversal strategies. To address these challenges, we introduce two core innovations: (1) an efficient knowledge graph construction pipeline that leverages dependency parsing to achieve $94\%$ of LLM-based performance ($61.87\%$ vs. $65.83\%$) while significantly reducing costs and improving scalability; and (2) a hybrid retrieval strategy that fuses vector similarity with graph traversal using Reciprocal Rank Fusion (RRF), maintaining separate embeddings for entities, chunks, and relations to enable multi-granular matching. We evaluate our framework on two enterprise datasets focused on legacy code migration and demonstrate improvements of up to $15\%$ and $4.35\%$ over vanilla vector retrieval baselines using LLM-as-Judge evaluation metrics. These results validate the feasibility of deploying GraphRAG in production enterprise environments, demonstrating that careful engineering of classical NLP techniques can match modern LLM-based approaches while enabling practical, cost-effective, and domain-adaptable retrieval-augmented reasoning at scale.

\end{abstract}

\section{Introduction}
Retrieval-Augmented Generation (RAG) has emerged as a practical framework for enhancing LLMs by grounding their outputs in external knowledge sources. In a standard RAG pipeline, a user query triggers the retrieval of semantically relevant passages from a document corpus using dense-vector retrieval. These retrieved passages are then fed to an LLM as contextual input, anchoring its responses in factual content. This architecture helps reduce hallucinations and enables the model to stay current with evolving information without the need for expensive model retraining~\cite{lewis2020retrieval}. In enterprise settings, RAG allows organizations to integrate proprietary data so that generated responses align with the latest domain-specific knowledge~\cite{gao2023retrieval}.

Modern enterprise resource planning (ERP) systems - used for finance, procurement, HR, and manufacturing - generate vast volumes of structured and unstructured data across interconnected modules. Enterprise queries often involve reasoning over configuration rules, transactional dependencies, change logs, and migration notes or cookbooks that are distributed across documents and systems in modern ERP systems. For example, assessing the impact of a custom code migration in S/4HANA \footnote{S/4HANA is an in-memory databse that the next-generation ERP system runs on.} may require linking legacy Advanced Business Application Programming (ABAP) \footnote{ABAP is a high-level programming language within the ERP ecosystem.} functions with deprecation reports, compatibility matrices, and policy guidelines. Traditional RAG systems treat documents as isolated units, limiting relational reasoning capabilities. Graph-based retrieval addresses this limitation by modeling entity relationships and enabling structure-aware context selection, making it well-suited for enterprise applications where understanding dependencies between components and processes is critical.

However, deploying GraphRAG in enterprise settings introduces two core challenges:
\begin{enumerate}
    \item \textbf{Computational cost of graph construction.} Building a knowledge graph (KG) at enterprise scale requires large-scale entity and relation extraction. When this process relies on LLMs or heavyweight NLP pipelines, it incurs significant GPU costs, leading to high latency and limited refresh frequency for dynamic content.
    \item \textbf{Retrieval latency and scalability.} Querying large graphs for relevant subgraphs introduces significant latency. Complex traversal and ranking operations struggle to meet real-time performance requirements at scale, even with optimized databases (DBs).
\end{enumerate}

In this paper, we propose a GraphRAG framework for enterprise-scale deployment with three key contributions: 
\begin{enumerate}
    \item \textbf{Efficient Knowledge Graph Construction.} A pipeline using dependency parsing that achieves competitive performance while reducing reliance on expensive LLM-based extraction.
    \item \textbf{Hybrid Retrieval Strategy.} Combining vector similarity with graph traversal using RRF~\cite{cormack2009reciprocal} to improve retrieval effectiveness.
    \item \textbf{Real-World Legacy Code Migration Application.} First application of GraphRAG to enterprise legacy code migration, demonstrating significant improvements over dense retrieval baselines.
\end{enumerate}
These contributions enable explainable, accurate, and scalable retrieval-augmented reasoning in complex enterprise environments.

\section{Related Work}
RAG combines dense-vector retrieval with language models (LM) to ground generation in external knowledge~\cite{lewis2020retrieval}. While effective for simple queries, traditional RAG systems treat documents as isolated units, limiting their effectiveness for relational reasoning over structured knowledge~\cite{barnett2024seven,bruckhaus2024rag}. This has motivated the development of graph-based approaches that explicitly model entity relationships.

To address these gaps, the GraphRAG paradigm was introduced, embedding a structured knowledge graph (KG) between the retrieval and generation stages~\cite{han2024retrieval}. GraphRAG~\cite{edge2024local} pioneered the integration of KGs into RAG by constructing entity-relation graphs from retrieved passages and organizing them into semantic communities through LLM-based summarization. This approach demonstrated significant improvements in multi-hop reasoning but incurs substantial computational costs due to extensive LLM usage during both construction and query-time summarization.
Recent work has focused on improving GraphRAG efficiency. LightRAG~\cite{guo2025lightragsimplefastretrievalaugmented} introduces dual-level entity-relation indexing to accelerate retrieval, while HippoRAG~\cite{jimenez2024hipporag} employs Personalized PageRank for memory-inspired graph traversal. FastGraphRAG~\cite{fastgraphrag2025} proposes optimizations for faster subgraph extraction. Most recently, SubGCache~\cite{zhu2025subgcache} addresses query-time latency through subgraph-level key-value (KV) caching to reduce redundant LLM inference during retrieval.
However, all these systems rely on LLM-based KG construction, which presents a fundamental scalability bottleneck for large enterprise corpora. While query-time optimizations like SubGCache improve retrieval efficiency, the upstream construction cost remains prohibitive for dynamic, large-scale deployments.

In contrast to prior work, we address the construction bottleneck by demonstrating that dependency-based extraction—leveraging classical NLP techniques—achieves 94\% of LLM-based performance while significantly reducing computational costs. We further introduce a hybrid retrieval strategy combining vector similarity with efficient graph traversal via RRF, maintaining separate embeddings for entities, chunks, and relations. Unlike previous GraphRAG systems that focus primarily on reasoning capabilities or query-time efficiency, our framework addresses both construction scalability and retrieval effectiveness, enabling practical GraphRAG deployment in cost-sensitive enterprise environments. We validate our approach on real-world legacy code migration tasks, demonstrating the first application of GraphRAG to this domain.

\section{Methodology}
Our GraphRAG framework comprises two core components designed for scalable enterprise deployment:
\begin{enumerate}
    \item Flexible \textbf {KG Construction}, supporting both dependency-based and LLM-based extraction modes, enabling cost-accuracy trade-offs based on deployment requirements
    \item \textbf{Hybrid Graph Retrieval}, combining efficient graph traversal with vector-based ranking to retrieve high-recall, semantically relevant contexts.
\end{enumerate}

\subsection{Knowledge Graph Construction}
We support two interchangeable construction pipelines: a dependency-based approach that leverages linguistic structure for fast, cost-effective extraction, and a LLM-based approach that achieves higher accuracy on smaller datasets. Both produce entity-relation graphs stored in a unified backend graph DB for downstream retrieval. 

\begin{figure}[htbp]
\centering
\includegraphics[width=0.8\textwidth]{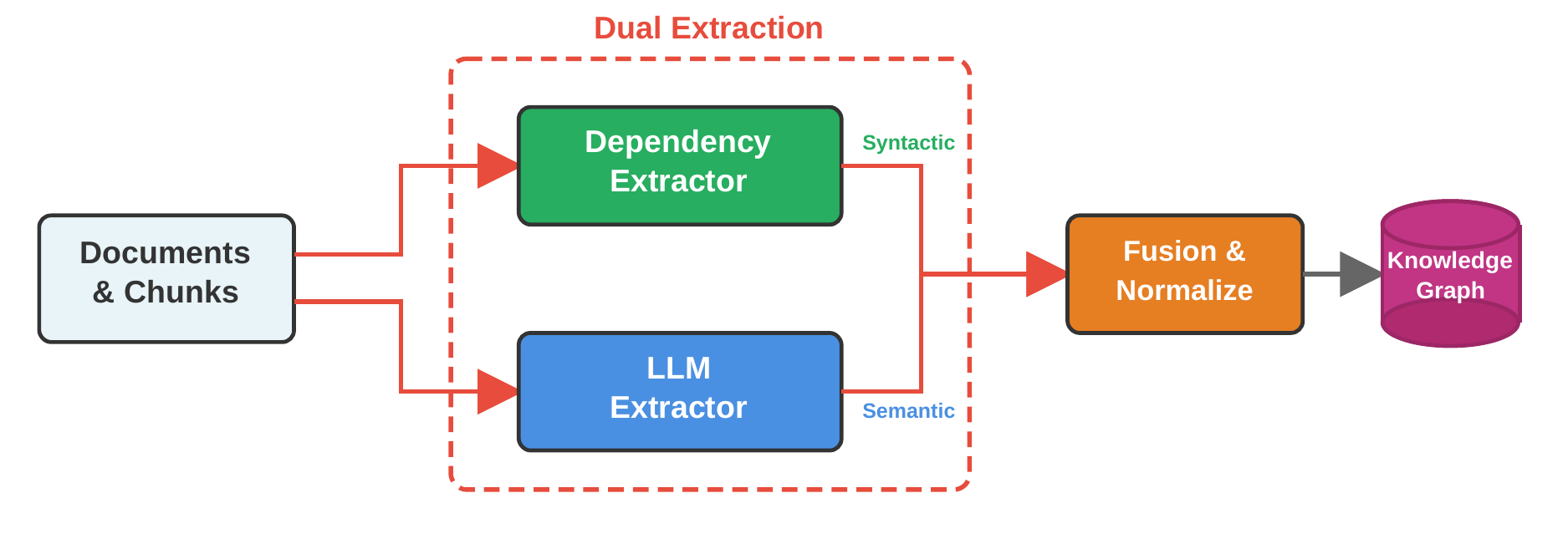}
\caption{Dual Extraction Architecture for Knowledge Graph Construction}
\label{fig:kg_architecture}
\end{figure}

Figure~\ref{fig:kg_architecture} illustrates our dual knowledge graph construction pipeline, which supports both an LLM-based extraction path, a lightweight dependency-parser-based alternative, and a combination of the two. Input documents pass through a series of preprocessing and filtering stages before triples are extracted, normalized, and materialized in the target graph store.
\vspace{-6mm}
\subsubsection{Preprocessing Pipeline}
Input documents arrive in diverse formats (PDF, HTML, XLSX, CSV) and undergo standardized preprocessing. We parse documents using Docling\footnote{\url{https://github.com/docling-project/docling}}  to extract text while preserving structural metadata, then apply hierarchical chunking~\cite{hearst-1997-text} that respects discourse boundaries by splitting at section headers. When a section exceeds $2048$ characters, we apply recursive character-level splitting with a $200$-character overlap. Each chunk is segmented into sentences using SpaCy~\footnote{\url{https://spacy.io/}}, and we filter sentences lacking verb phrases to reduce downstream processing overhead—an optimization that significantly improves efficiency for large corpora.
\vspace{-3mm}
\subsubsection{Dependency-Based Triple Extraction} 
\begin{figure}[h!]
  \centering
  \includegraphics[width=0.8\linewidth]{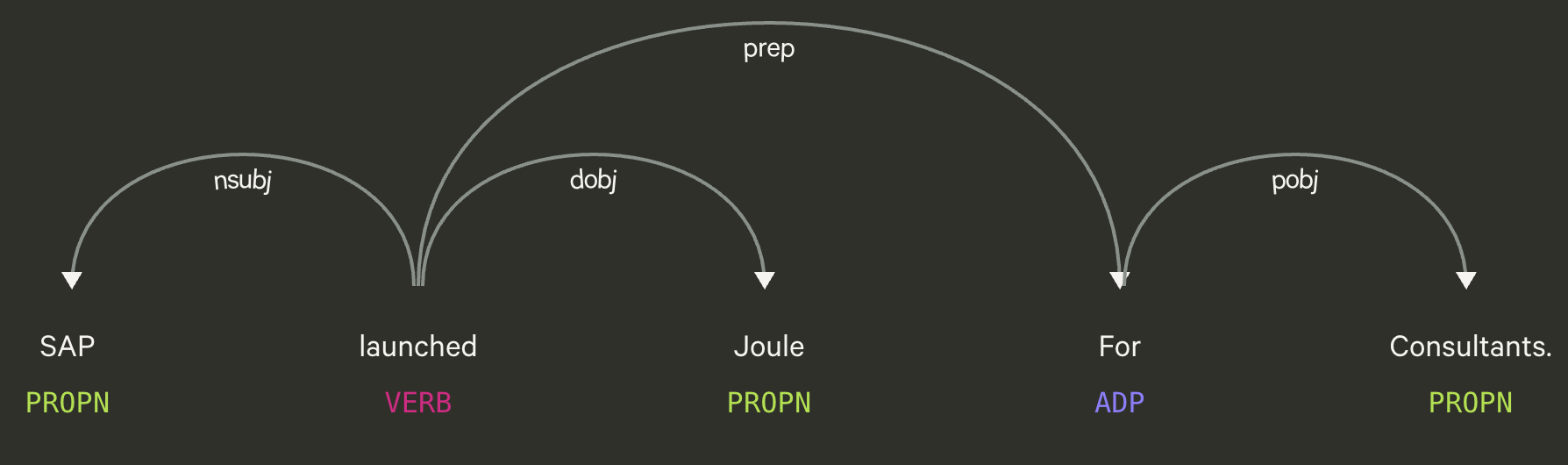}
  \caption{SpaCy Generated Parse Tree}
  \label{fig:spacy-tree}
\end{figure}

We draw upon dependency grammar theory~\cite{de-marneffe-etal-2021-universal}, which posits that a sentence’s syntactic structure can be represented as a graph of binary head–dependent relations. A core contribution of our work in this paper demonstrates that dependency parsing can achieve competitive knowledge extraction performance while maintaining enterprise-scale efficiency. We leverage SpaCy's dependency parser to extract entity-relation triples directly from syntactic structure. For example, for the sentence \textit{"SAP launched Joule for Consultants"}, the dependency parser produces a tree structure (Figure~\ref{fig:spacy-tree}) identifying \textit{"launched"} as the root verb with \textit{"SAP"} as subject (nsubj), \textit{"Joule"} as direct object (dobj), and \textit{"Consultants"} as prepositional object. From this structure, we extract triples: \textit{("SAP", "launched", "Joule")} and \textit{("Joule", "for", "Consultants")}. Ideally, \textit{"Joule for Consultants"} should be recognized as an named entity and merged as one token, and then it should be parsed as one direct object of the predicate \textit{"launched"}.

Algorithm~\ref{alg:dep-kg-coref} formalizes our dependency extraction logic. We construct a custom SpaCy processing pipeline that incorporates:

\begin{itemize}
\item \textbf{Passive voice handling} to normalize active/passive constructions
\item \textbf{Phrasal merging} to capture multi-token entities (e.g., \textit{"Supplier management"})
\item \textbf{Coreference resolution} to map pronouns and mentions to canonical entities
\item \textbf{Dependency triple extraction} identifying subject-verb-object patterns
\item \textbf{Linear extraction heuristics} to capture relationships missed by dependency parsing
\end{itemize}

\begin{algorithm}[h!]\scriptsize
\caption{\scriptsize Dependency-Based Knowledge Graph Construction with Coreference\protect\footnotemark}
\label{alg:dep-kg-coref}
\begin{algorithmic}[1]
\State $\texttt{nlp} \gets \textsc{BuildPipeline()}$
\Comment{includes: customized tokenizer, passive \& phrasal merges, entity merging, and coreference $+$ span resolver}
\State $\texttt{Entities} \gets \emptyset;\ \texttt{Relations} \gets \emptyset$
\For{\textbf{each} \texttt{text} \textbf{in} \texttt{Texts}}
  \State $\texttt{doc} \gets \texttt{nlp(text)}$
  \State $\texttt{dep\_triples} \gets \textsc{ExtractDependencyTriples(doc)}$
  \State $\texttt{linear\_triples} \gets \textsc{LinearExtractor(doc)}$
  \State $\texttt{all\_triples} \gets \texttt{dep\_triples} \cup \texttt{linear\_triples}$
  \State $\texttt{coref\_map} \gets \textsc{BuildCorefMap(doc)}$
  \State $\texttt{resolved\_triples} \gets \{(\texttt{coref\_map.get}(h,h), r, \texttt{coref\_map.get}(o,o))\ |\ (h,r,o)\in\texttt{all\_triples}\}$
  \State $\texttt{filtered\_triples} \gets \{(\textsc{Normalize}(h), r, \textsc{Normalize}(o))\ |\ (h,r,o)\in\texttt{resolved\_triples},$
  \Statex \hspace{2em} $\texttt{len}(h)\ge2 \wedge \texttt{len}(o)\ge2 \wedge h.\texttt{lower()}\notin\texttt{WORD\_FILTER} \wedge o.\texttt{lower()}\notin\texttt{WORD\_FILTER}\}$
  \For{\textbf{each} $(h,r,o)$ \textbf{in} \texttt{filtered\_triples}}
    \State $\texttt{Entities} \gets \texttt{Entities} \cup \{(\textsc{Id}(h),\ \textit{name}=h,\ \textit{type}=\text{"Concept"}),$
    \Statex \hspace{3em} $(\textsc{Id}(o),\ \textit{name}=o,\ \textit{type}=\text{"Concept"})\}$
    \State $\texttt{confidence} \gets \textsc{ScoreRelation}(r,h,o)$ \Comment{optional heuristic}
    \State $\texttt{Relations} \gets \texttt{Relations} \cup \{(\textit{head}=h,\ \textit{relation}=r,\ \textit{tail}=o,\ \textit{confidence}=\texttt{confidence})\}$
  \EndFor
\EndFor
\State \Return $\texttt{Entities}, \texttt{Relations}$
\end{algorithmic}
\end{algorithm}
\footnotetext{Notation: \textbf{Bold} denotes control flow keywords, \textsc{SmallCaps} denotes function calls, and \texttt{typewriter} denotes variables.}

For each document, we extract dependency-based and linear triples, resolve coreferences to canonical forms, and apply normalization, filtering short entities ($< 2$ characters), removing stopwords, and standardizing entity names for graph DB compatibility. Each relation is assigned a confidence score based on syntactic features.
A key advantage of this approach is domain agnosticism—the method requires no domain-specific training or customization, making it directly applicable across diverse enterprise contexts. As we demonstrate in Section \ref{sec:experiments}, this dependency-based extraction achieves $94\%$ of LLM-based performance ($61.87\%$ vs. $65.83\%$) while processing documents orders of magnitude faster and at significantly lower cost.
\vspace{-6mm}
\subsubsection{LLM-Based Extraction} For critical document collections or text with complex ambiguity where maximum accuracy is required, our framework supports LLM-based extraction using \texttt{GPT}\footnote{https://platform.openai.com/docs/models} family of models with few-shot prompting. Users can select extraction mode based on their cost-performance requirements and document characteristics, enabling practical deployment across varying enterprise scenarios.
\subsubsection{Graph Storage} Extracted entities and relations are stored in a graph DB (i.e. iGraph ~\cite{igraph}) with vector embeddings generated for each entity, chunk, and relation using OpenAI's \texttt{text-embedding-3-large}\footnote{https://platform.openai.com/docs/models/text-embedding-3-large} model. These embeddings enable hybrid retrieval that combines graph structure with semantic similarity, as described in Section \ref{sec:eff_hybrid_graph}.
\vspace{-8mm}
\subsection{Efficient Hybrid Graph Retrieval}
\label{sec:eff_hybrid_graph}
\vspace{-10mm}
At query time, we employ a cascaded retrieval strategy that combines efficient graph traversal with vector-based ranking. First, we conduct a high-recall one-hop graph traversal to identify candidate nodes. Second, based on candidate nodes, we perform 1-hop traversal to retrieve neighbors to obtain subgraphs. Next, we apply a dense vector-based re-ranking step using embeddings and cosine similarity to refine the result set. The selected subgraph, along with relevant source text chunks and extracted query entities, is then passed to an LLM to generate response. Our retrieval approach aligns with the classical cascaded architecture in information retrieval (IR), where an initial recall-oriented stage (e.g., BM25 or dense vector search) is followed by a precision-oriented neural re-ranker~\cite{mogotsi2010christopher,nogueira2020passagererankingbert,adjali2024multi}. Our one-hop traversal effectively retrieves semantically related nodes while keeping the candidate set size tractable—crucial for scaling to large enterprise graphs.

\begin{figure}[H]
    \centering
    \includegraphics[width=\columnwidth]{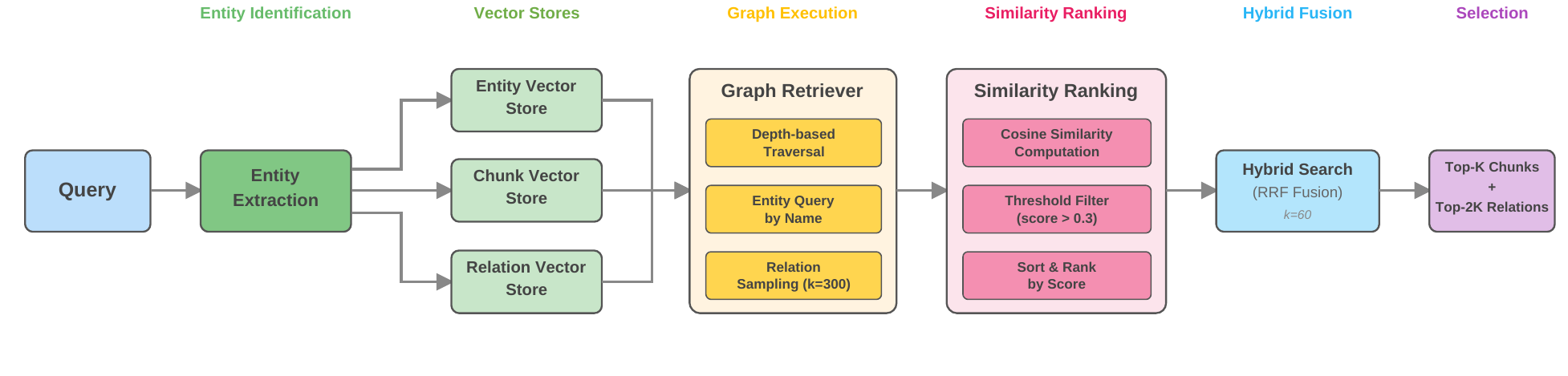}
    \caption{GraphRAG Retrieval Architecture}
    \label{fig:graphrag_retrieval}
\end{figure}

Figure~\ref{fig:graphrag_retrieval} illustrates the major components in our indexing and retrieval pipeline. During indexing, the KG is stored in both vector DB and graph DB. For our experiments, we use the open-source library Milvus~\cite{2021milvus} for storing embeddings and high-performance iGraph to store the graph in memory. Milvus stores nodes, chunks and relation embeddings for fast similarity lookup at query time, and iGraph stores nodes and edges for fast traversal. The following sections describe the major components for the retrieval processes. Algorithm~\ref{alg:graphrag} provides the complete retrieval procedure.

\begin{algorithm}[H]\scriptsize
\caption{\scriptsize GraphRAG Hybrid Retrieval}
\label{alg:graphrag}
\begin{algorithmic}[1]
\State $E_{\text{seed}} \gets \textsc{NounPhraseExtraction}(Q) \cup \textsc{VectorSearch}(Q, \mathcal{V}, k{=}5)$
\Comment{Extract seed entities via noun phrases and top-5 vector similarity}

\State $R \gets \emptyset;\ Ch \gets \emptyset$
\For{\textbf{each} $e$ \textbf{in} $E_{\text{seed}}$}
  \State $v \gets \textsc{ExactMatch}(e, G)$ \Comment{Case-insensitive match in graph $G$}
  \If{$v$ found}
    \State $N \gets \textsc{Get1HopNeighbors}(v, G)$
    \State $\texttt{sampled} \gets \textsc{SampleRelations}(N, k)$ \Comment{$k{=}100$ for small/medium, $k{=}200$ for large graphs}
    \State $R \gets R \cup \texttt{sampled.relations}$
    \State $Ch \gets Ch \cup \texttt{sampled.chunks}$
  \EndIf
\EndFor

\State $L_{\text{graph}}^{(ch)} \gets \textsc{RankBySimilarity}(Ch, Q, \mathcal{V})$ \Comment{Rank chunks by cosine similarity}
\State $L_{\text{graph}}^{(r)} \gets \textsc{RankBySimilarity}(R, Q, \mathcal{V})$ \Comment{Rank relations by cosine similarity}
\State $L_{\text{vector}} \gets \textsc{DenseVectorSearch}(Q, \mathcal{V})$ \Comment{Pure vector search on all chunks}

\State $L_{\text{fused}} \gets \textsc{RecipRankFusion}(L_{\text{graph}}^{(ch)}, L_{\text{vector}}, k{=}60)$ \Comment{Hybrid fusion for chunks}

\State $Ch_{\text{top-k}} \gets \textsc{SelectTop}(L_{\text{fused}}, k)$
\State $R_{\text{top-2k}} \gets \textsc{SelectTop}(L_{\text{graph}}^{(r)}, 2k)$
\State $\mathcal{C} \gets \{Ch_{\text{top-k}}, R_{\text{top-2k}}, E_{\text{seed}}\}$

\State \Return $\mathcal{C}$
\end{algorithmic}
\end{algorithm}
\vspace{-4mm}
\subsubsection{Query Entity Identification}
In contrast to other GraphRAG methods~\cite{guo2025lightragsimplefastretrievalaugmented,jimenez2024hipporag,fastgraphrag2025} that solely rely on LLMs for entity identification, we employ an optimized variant of SpaCy's noun phrase extractor we developed to efficiently pinpoint key concepts within the query. Additionally, we conduct a similarity search between the full query and node embeddings to retrieve the $\text{top-}k$, where $k=5$ relevant nodes from the graph. The entities obtained from both approaches are then merged and used as seed nodes for relation extraction. We maintain separate vector embeddings for entities, chunks, and relations to enable multi-granular similarity matching during retrieval.
\vspace{-4mm}
\subsubsection{Graph Query Execution}
Starting from seed query nodes, we use case insensitive exact match to query the graph for relevant relations. Once a node is matched with a query node, it performs $1$-hop traversal of all neighbors and filtered by a neighbor controlling parameter $random\_k\_relations$. For small to medium size graph, $random\_k\_relations=100$ is sufficient, for larger ones, we set the parameter to $200$ akin to Yasunaga et al~\cite{yasunaga2022deepbidirectionallanguageknowledgegraph}. This yields a candidate set of entity-to-entity relations and entity-to-chunk associations.
\vspace{-4mm}
\subsubsection{Relevance Ranking and Context Selection}
Once the candidate relations are obtained through case insensitive exact match and graph traversal, they are split into two groups: entity-to-entity relations and entity-to-chunk relations. Both chunk and relation embeddings are retrieved from the vector DB, which are then used to compute cosine similarity with the query. Chunks and relations are then sorted by similarity scores, and $\text{top-}k$ chunks and $\text{top-}k *2$ relations are returned. In selecting $\text{top-}k$ chunks, our GraphRAG approach employs RRF to combine results from dense vector search and 1-hop graph traversal, balancing semantic similarity with structural entity relationships for improved context selection. This hybrid strategy balances structural relationships captured by the graph with semantic understanding from embeddings. 
\vspace{-4mm}
\subsubsection{Context Integration with LLM}
Once the $\text{top-}k$ chunks and $\text{top-}k *2$ relations are produced, we send them along with query entities as the context for LLM to consume and generate answers. Context is a dictionary with three keys: $Context = \{"chunks":chunk\_list, "relations": relation\_list, "entity": entity\_list\}$, which provides a much richer context than standard RAG alone.

\section{Experiments}
\label{sec:experiments}

\subsection{Datasets}
We evaluate our framework on Custom Code Migration (CCM) \footnote{Custom code migration is the process of identifying, analyzing, updating, and transferring custom-built ABAP components from an old system to a new one.}, a real-world enterprise use-case requiring technical understanding of ABAP code migration and system evolution. 

CCM resource corpus consists of $550$ PDF documents, including Cookbooks and Notes related to ABAP code migration. These documents are preprocessed into approximately $2000$ text chunks, each with a length of $2048$ characters and an overlap of $200$ characters. This processed corpus serves as the foundation for both KG construction and dense vector representation in our experiments.  

Two test datasets are designed to evaluate different aspects of the system. CCM Chat includes $150$ question-answer pairs focused on code migration topics, including error analysis, implementation differences, and best practices for transitioning from legacy to S/4HANA systems. CCM Code Proposal comprises $200$ legacy code examples, each containing the legacy code alongside the migrated version.

Using the CCM resource corpus, our method yields a KG with $39155$ nodes, $47613$ entity-to-entity relations, $63681$ entity-to-chunk relation, resulting in an average node degree of $1.52$ and a highest degree of $236$. This relatively sparse structure reflects the technical, domain-specific nature of the corpus while maintaining sufficient connectivity for effective graph traversal.

\subsection{Evaluation Methodology} 

\subsubsection{CCM Chat Evaluation}
We employ two complementary evaluation techniques: \begin{inparaenum}[(i)]
    \item \textit{\textbf{Semantic Alignment Score}} and \item \textit{\textbf{RAGAS Score}}. 
\end{inparaenum}

\textit{\textbf{Semantic Alignment Score}} uses an LLM-based classifier to compare the generated response against a reference (ground truth) answer. The LLM is prompted to assign a discrete semantic coverage score: \begin{inparaenum}[(i)] \item \textbf{$0$} if the answer fails to cover any part of the ground truth, \item \textbf{$0.5$} if it partially captures key information, and \item \textbf{$1$} if it fully aligns with the reference answer. \end{inparaenum} An overall performance metric is computed as a weighted average:

$$\text{Semantic Alignment Score} = (0.5 \times P_{0.5} + 1.0 \times P_{1.0}) \times 100\%$$
 where $P_{0.5}$ and $P_{1.0}$ represent the proportions of responses assigned scores of $0.5$ and $1.0$, respectively. This metric provides a nuanced assessment that accounts for partial correctness, common in technical question answering where responses may capture some but not all relevant information.

\textit{\textbf{RAGAS Score}}~\cite{es2025ragasautomatedevaluationretrieval} assesses both the retrieval quality and generative accuracy using three metrics: \begin{inparaenum}[(i)] 
\item \textit{Context Precision}, which measures the proportion of retrieved chunks that are relevant to the question;
\item \textit{Faithfulness}, which quantifies how much of the generated answer is grounded in the retrieved content; and 
\item \textit{Answer Relevancy}, which evaluates how directly the generated response addresses the original query. This is computed by generating follow-up questions based on the response and comparing their cosine similarity to the original query—a higher similarity indicates stronger relevance.
\end{inparaenum}

\subsubsection{CCM Code Proposal Evaluation}
We adopt an \textit{LLM-as-a-Judge} framework to systematically compare generated migration code with ground truth. The evaluation uses a structured two-stage prompt design: \begin{inparaenum}[(i)] \item a system prompt that defines the evaluation task, instructing the judge to compare dense vector and graph-based generated code against human-created ground truth across five technical dimensions (as listed below), and \item a user prompt that provides all four code inputs (legacy code, ground truth, dense vector output, graph output) and requests structured output. \end{inparaenum} Each evaluation instance includes the original legacy code, the reference migrated version created by human experts, and two system-generated outputs—one produced using dense vector retrieval and the other via graph-based retrieval. The judge outputs both a winner selection and detailed scoring breakdowns in JSON format as shown in\footnote{Complete evaluation prompts and code examples are available in our code repository at \url{https://anonymous.4open.science/r/graphrag-pakdd2026-evaluation-3FBC} (anonymized for review).}.

Evaluation proceeds in two stages: \begin{inparaenum}[(i)]
\item \textit{\textbf{Pairwise Comparison}}, where the LLM selects the more accurate candidate based on similarity to the ground truth; and 
\item \textit{\textbf{Rubric Scoring}}, where each output is rated (1-5 scale) across five criteria: \textit{Syntax Correctness}, \textit{Logical Correctness}, \textit{S/4HANA Compatibility}, \textit{Optimization and Efficiency}, and \textit{Readability} . 
\end{inparaenum}
\vspace{-10mm}
\subsection{Results and Analysis}
\vspace{-6mm}
\subsubsection{System Performance Result On CCM Chat}
In Table~\ref{tab:ragas_eval} and Table~\ref{tab:llm_judge_eval}, both variants of GraphRAG, one using \texttt{GPT-4o} and the other using dependency graph as triplet creation model, show at least $12\%$ improvement in context precision score compared to dense vector retrieval. In terms of semantic alignment (abbreviated as No Cov., Partial Cov., and Full Cov. in Table~\ref{tab:llm_judge_eval}), the \textit{No Cov.} rate is reduced by $32\%$ for both variants, while the \textit{Full Cov.} (complete alignment with ground truth) rate increases by at least $19\%$. Table \ref{tab:llm_judge_eval} shows that both GraphRAG variants substantially improve answer completeness, with weighted averages of $65.83\%$ and $61.87\%$ compared to $50.80\%$ for dense retrieval. This improvement primarily stems from increased full-coverage responses ($58.99\%$ and $51.08\%$ vs. $42.88\%$) and reduced no-coverage failures ($27.34\%$ vs. $40.29\%$), demonstrating that graph-structured retrieval better captures complete entity-grounded context for technical queries.

Notably, the dependency graph-based GraphRAG model retains $94\%$ of the \texttt{GPT-4o} variant’s performance in context precision. It achieves comparable results in \textit{No Cov.} and reaches $86.6\%$ of the \texttt{GPT-4o} variant’s performance in \textit{Full Cov.} highlighting its strong performance with a lighter KG construction pipeline.

\begingroup
\setlength{\intextsep}{6pt}      
\setlength{\textfloatsep}{6pt}   
\setlength{\floatsep}{6pt}       
\setlength{\abovecaptionskip}{2pt}
\setlength{\belowcaptionskip}{2pt}

\FloatBarrier  

\begin{table}[H]
    \caption{RAGAS evaluation on CCM chat}
    \label{tab:ragas_eval}
    \centering
    {\footnotesize
    \resizebox{0.85\textwidth}{!}{ 
    \begingroup
    \setlength{\tabcolsep}{10pt}
    \renewcommand{\arraystretch}{1.1}
    \begin{tabular}{l@{\hspace{6pt}}c@{\hspace{6pt}}c@{\hspace{6pt}}c@{\hspace{6pt}}c}
        \toprule
        \textbf{Method} &
        \makecell{\textbf{Context}\\\textbf{Precision}} &
        \textbf{Faithfulness} &
        \makecell{\textbf{Answer}\\\textbf{Relevancy}} &
        \textbf{Avg.} \\
        \midrule
        Dense Vector (\texttt{ada-002}) & 54.35\% & 77.18\% & 82.92\% & 71.48\% \\
        GraphRAG (\texttt{GPT-4o}) & 63.82\% & 74.24\% & 89.43\% & 75.83\% \\
        \textbf{GraphRAG (Dependency)} & \textbf{61.07\%} & \textbf{72.76\%} & \textbf{90.97\%} & \textbf{74.93\%} \\
        \bottomrule
    \end{tabular}
    \endgroup
    }
    }
\end{table}

\begin{table}[H]
    \caption{Semantic Alignment evaluation on CCM chat}
    \label{tab:llm_judge_eval}
    \centering
    {\footnotesize
    \resizebox{0.85\textwidth}{!}{ 
    \begingroup
    \setlength{\tabcolsep}{10pt}
    \renewcommand{\arraystretch}{1.1}
    \begin{tabular}{l@{\hspace{6pt}}c@{\hspace{6pt}}c@{\hspace{6pt}}c@{\hspace{6pt}}c}
        \toprule
        \textbf{Method} &
        \makecell{\textbf{No}\\\textbf{Cov. (0)}} &
        \makecell{\textbf{Partial}\\\textbf{Cov. (0.5)}} &
        \makecell{\textbf{Full}\\\textbf{Cov. (1)}} &
        \makecell{\textbf{Weighted}\\\textbf{Avg.}} \\
        \midrule
        Dense Vector (\texttt{ada-002}) & 40.29\% & 15.85\% & 42.88\% & 50.80\% \\
        GraphRAG (\texttt{GPT-4o}) & 27.34\% & 13.67\% & 58.99\% & 65.83\% \\
        \textbf{GraphRAG (Dependency)} & \textbf{27.34\%} & \textbf{21.58\%} & \textbf{51.08\%} & \textbf{61.87\%} \\
        \bottomrule
    \end{tabular}
    \endgroup
    }
    }
\end{table}

\FloatBarrier  
\endgroup

\subsubsection{\textbf{System Performance Result on CCM Code Proposal}} In Table~\ref{tab:ccm_code_gpt4o} and Table~\ref{tab:ccm_code_dependency} on \textit{CCM Code Proposal} dataset, both GraphRAG variants outperform dense retrieval in terms of winning rate and average score, where average score measures an average across all five evaluation criteria and winning rate is defined as the ratio of cases in which a model’s response is preferred over the baseline according to the LLM-as-a-Judge assessment. The GraphRAG system leveraging dependency parsing achieves performance on par with \texttt{GPT-4o} variant, indicating that dependency graph-based GraphRAG is a strong alternative to LLM-based triplet extraction in retrieval tasks on this dataset.

\begingroup
\setlength{\intextsep}{6pt}      
\setlength{\textfloatsep}{6pt}   
\setlength{\floatsep}{6pt}       
\setlength{\abovecaptionskip}{2pt}
\setlength{\belowcaptionskip}{2pt}

\FloatBarrier  

\begin{table}[H]
  \caption{LLM-as-a-Judge on CCM Code Proposal (\texttt{GPT-4o}-based)}
  \label{tab:ccm_code_gpt4o}
  \centering
  {\footnotesize
  \resizebox{0.65\textwidth}{!}{ 
  \begingroup
  \setlength{\tabcolsep}{10pt}
  \renewcommand{\arraystretch}{1.1}
  \begin{tabular}{l@{\hspace{6pt}}c@{\hspace{6pt}}c}
    \toprule
    \textbf{Method} &
    \makecell{\textbf{Winning}\\\textbf{Rate}} &
    \makecell{\textbf{Avg.}\\\textbf{Score (1--5)}} \\
    \midrule
    Dense Vector (\texttt{ada-002}) & 23\% & 3.48 \\
    \textbf{GraphRAG (\texttt{GPT-4o})} & \textbf{77\%} & \textbf{4.04} \\
    \bottomrule
  \end{tabular}
  \endgroup
  }
  }
\end{table}

\begin{table}[H]
  \caption{LLM-as-a-Judge on CCM Code Proposal (Dependency Graph-based)}
  \label{tab:ccm_code_dependency}
  \centering
  {\footnotesize
  \resizebox{0.65\textwidth}{!}{ 
  \begingroup
  \setlength{\tabcolsep}{10pt}
  \renewcommand{\arraystretch}{1.1}
  \begin{tabular}{l@{\hspace{6pt}}c@{\hspace{6pt}}c}
    \toprule
    \textbf{Method} &
    \makecell{\textbf{Winning}\\\textbf{Rate}} &
    \makecell{\textbf{Avg.}\\\textbf{Score (1--5)}} \\
    \midrule
    Dense Vector (\texttt{ada-002}) & 21.5\% & 3.43 \\
    \textbf{GraphRAG (Dependency)} & \textbf{78.5\%} & \textbf{4.03} \\
    \bottomrule
  \end{tabular}
  \endgroup
  }
  }
\end{table}

\FloatBarrier  
\endgroup

\subsubsection{Qualitative Insight} Our method is effective in retrieving content tied to key entities. For example, in \textit{CCM chat}, given question \textit{"How do I handle custom code that references \textit{VBBS}\footnote{VBBS is a legacy database table that stores summarized sales requirement totals.} after the S/4HANA conversion?"}, dense retriever fails to extract content that specifically address VBBS table. In contrast, GraphRAG selects content explicitly mentioning \textit{"VBBS"} and includes relevance sentences: \textit{“… If the VBBS is used in customer code, … The solution is to create a view on VBBE \footnote{VBBE is a database table that stores individual, detailed sales requirments for Material Requirements Planning (MRP).}”}. This illustrates our system's strength in extracting semantically focused context.

In \textit{CCM Code Proposal}, GraphRAG exhibits structured multi-entity reasoning that dense retrieval lacks. Dense vector retrieval correctly updates the transaction code but fails to migrate the corresponding screen references, retaining the obsolete \texttt{SAPMM03S/RM03S} structures that cause runtime errors in S/4HANA. GraphRAG, by traversing entity relationships in the KG (linking transactions to their required screen structures), correctly identifies that \texttt{MSC3N} requires \texttt{SAPLMGMM/RMMG1} and performs complete migration of both components. Please refer to prompts and evaluation samples\footnote{Complete evaluation prompts and code examples: \url{https://anonymous.4open.science/r/graphrag-pakdd2026-evaluation-3FBC}} for more details. 

\section{Conclusion, Limitation and Future Work}
In this work, we present a scalable method for constructing enterprise-grade graph-based GraphRAG systems from unstructured text. To address key scalability challenges in real-world enterprise environments, our approach centers on two core components: \begin{inparaenum}[(i)]
    \item KG construction using efficient dependency parsing to complement LLM-based approach, and \item lightweight, hybrid subgraph retrieval to ensure low-latency query-time performance. \end{inparaenum} We validate our framework on two use cases, CCM Chat and CCM Code Proposal, and observe consistent performance improvements over a baseline RAG system. Notably, KGs generated using a robust, open-source dependency parser achieved performance comparable to \texttt{GPT-4o}, as measured by both LLM-as-a-Judge and RAGAS evaluation metrics.

Our approach offers a promising path for scaling GraphRAG systems by alleviating the bottleneck of sole dependence on LLMs for KG construction. Nevertheless, two limitations warrant future investigation. First, while dependency parsing provides a lightweight and scalable method for extracting knowledge triples, it may miss context-dependent or implicit relations not directly expressed in surface syntax. Second, although our method demonstrates strong performance in code migration domain, its generalizability to other settings remains an open question. Future work includes evaluating the approach on broader public benchmarks such as HotpotQA to assess its applicability beyond enterprise use cases. Additionally, investigating advanced graph traversal strategies beyond one-hop and integrating with recent query-time optimizations like SubGCache represent promising directions for further improving retrieval efficiency.

\section{GenAI Usage Disclosure}
We employed ChatGPT and Claude to assist in rephrasing certain sections of the paper for improved clarity. All core content, including research design, data analysis, and result interpretation, was conducted without the aid of generative AI tools.

\bibliographystyle{unsrt}
\bibliography{pakdd}    

\end{document}